\documentclass[runningheads]{llncs}
\usepackage{graphicx}
\usepackage{times}
\usepackage{epsfig}
\usepackage{amsmath}
\usepackage{amssymb}
\usepackage{booktabs} 
\usepackage{enumerate}
\usepackage{adjustbox}
\usepackage{indentfirst}
\usepackage{array}
\usepackage{multirow}
\usepackage{float}
\usepackage{hhline}
\usepackage{bm}
\usepackage{subfigure}
\usepackage{color}

\usepackage{fancyhdr}

\begin{document}

\title{Dual CNN Models for Unsupervised Monocular Depth Estimation}

\author{Vamshi Krishna Repala \and Shiv Ram Dubey}

\institute{Computer Vision Group, \\
Indian Institute of Information Technology, Sri City, Chittoor, Andhra Pradesh, India.\\
\email{vamshi.r14@iiits.in, srdubey@iiits.in}
}

\maketitle

\thispagestyle{fancy}
\fancyhf{}
\lhead{Accepted in 8th Pattern Recognition and Machine Intelligence Conference 2019}

\begin{abstract}
The unsupervised depth estimation is the recent trend by utilizing the binocular stereo images to get rid of depth map ground truth. In unsupervised depth computation, the disparity images are generated by training the CNN with an image reconstruction loss. In this paper, a dual CNN based model is presented for unsupervised depth estimation with 6 losses (DNM6) with individual CNN for each view to generate the corresponding disparity map. The proposed dual CNN model is also extended with 12 losses (DNM12) by utilizing the cross disparities. The presented DNM6 and DNM12 models are experimented over KITTI driving and Cityscapes urban database and compared with the recent state-of-the-art result of unsupervised depth estimation. 
The code is available at: https://github.com/ishmav16/Dual-CNN-Models-for-Unsupervised-Monocular-Depth-Estimation.

\end{abstract}

\section{Introduction}
The image based depth estimation of scene is a very active research area in the field of computer vision. 
The depth map from images can be estimated in various ways like structure from motion \cite{nister2005preemptive}, multi-view stereo \cite{seitz2006comparison}, monocular methods \cite{saxena2006learning}, single-image methods \cite{saxena20083}, etc. 
The deep learning and convolutional neural networks (CNNs) based methods perform outstanding in most of the problems of computer vision such as image classification \cite{alexnet}, facial micro-expression recognition \cite{reddy2019spontaneous}, face anti-spoofing \cite{nagpal2018performance}, hyper-spectral image classification \cite{roy2019hybridsn}, image-to-image transformation \cite{kancharagunta2019csgan}, colon cancer nuclei classification \cite{basha2018rccnet}, etc. 
Inspired from the success of deep learning, several researchers also tried to utilize the CNN for the depth prediction, specially in monocular imaging conditions. These approaches are classified mainly in three categories namely learning-based stereo \cite{zhou2016learning}, \cite{zbontar2016stereo}, supervised single view depth estimation \cite{eigen2014depth}, \cite{liu2016learning}, and unsupervised depth estimation \cite{garg2016unsupervised}, \cite{godard2017unsupervised}. 
The stereo image pairs and ground truth disparity data are needed in order to train the learning-based stereo models. In real scenario, creating such data is very difficult. Moreover, these methods generally create the artificial data which can not represent the real challenges appearing in natural images and depth maps. 
The supervised single view depth estimation methods also use ground truth depth to train the model. 
The main hurdle in supervised approaches is availability and creation of ground truth depth maps which is always not available in real applications.

The unsupervised depth estimation methods do not need any ground truth depth maps. Basically, they utilize the underlying theory of epipolar constraints \cite{hartley2003multiple}. Recently, Garg et al. used auto-encoder deep CNN to predict the inverse depth map (i.e. disparity) from left image \cite{garg2016unsupervised}. They computed a warp image (i.e. reconstructed left image) from disparity map and right image. Finally, the error between original and reconstructed left image is used as the loss to train the whole setup in unsupervised manner. This approach is further improved by Godard et al. by incorporating the left-right consistency \cite{godard2017unsupervised}. In left-right consistency, basically two depth maps (i.e. left and right) are generated using auto-encoder only from the left input image. The left input image is used with generated right depth map and the right image is used with generated left depth map to reconstruct the right and left images respectively. Zhou et al. \cite{zhou2017unsupervised} utilized the concepts of unsupervised image depth estimation proposed in \cite{eigen2014depth} and \cite{godard2017unsupervised} to tackle the monocular depth and camera motion estimation in unstructured video sequences in unsupervised learning framework. In one of the recent work, the 3D loss such as photometric quality of frame reconstructions is combined with 2D loss such as pixel-wise or gradient-based loss for learning the depth and ego-motion from monocular video in unsupervised manner \cite{mahjourian2018unsupervised}.

While the unsupervised based methods have gained the attention in recent times, there is still need of discovering better suited unsupervised networks and loss functions. Through this paper, we propose a dual CNN based model for unsupervised monocular image depth estimation by utilizing the 6 losses (DNM6). We also extend the dual CNN model with 12 losses and generate DNM12 architecture to improve the quality of depth maps. The appearance matching loss, disparity smoothness loss and left-right consistency loss are used in this paper. 
The rest of the paper is structured by presenting the proposed dual CNN models DNM6 and DNM12 in Section 2, the experimental results and analysis in Section 3, and the concluding remarks in Section 4.

\section{Proposed Methodology}

\subsection{Dual Network Model with 6 Losses (DNM6)}
\begin{figure*}[t!]
\centering
\includegraphics[width=\textwidth]{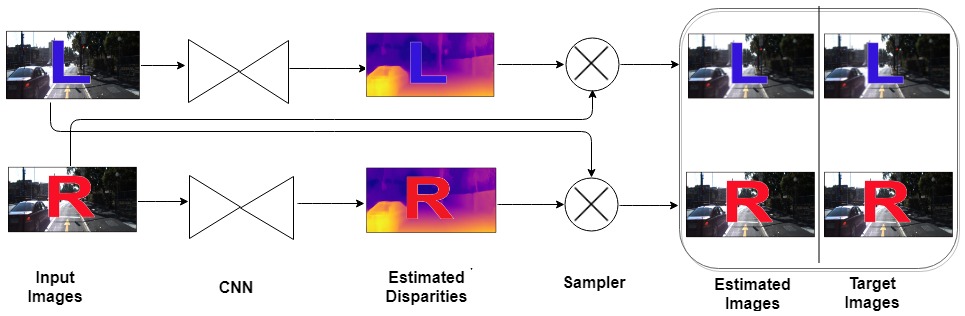}
\caption{Pictorial representation of proposed Dual Network Model with 6 Losses (DNM6)}
\label{fig:6loss}
\end{figure*}

The proposed idea of dual network model (DNM) using CNN is illustrated in Figure \ref{fig:6loss}. This model is based on the 6 losses, thus referred as the DNM6 model. The DNM6 model has two CNN one for each left and right images of stereo pair. During training, the left image $ I^{l} $ and right image $ I^{r} $ are considered as the inputs to the left CNN named as CNN-L and right CNN named as CNN-R respectively. The $ I_{i,j} $ refers to the $(i,j)^{th}$ co-ordinate of image $I$. It is assumed that both $ I^{l} $ and $ I^{r} $ images are captured in similar settings. Both CNN's are based on the auto-encoder algorithm and combined these two networks named as dual network. The CNN architecture (in both CNNs) is taken from the Godard et al. \cite{godard2017unsupervised}. The CNN-L predicts the left disparity map $ d^{l} $, whereas the CNN-R predicts the right disparity map $ d^{r} $. The $ d_{i,j} $ refers to disparity value at $(i,j)^{th}$ co-ordinate of disparity map $d$. In order to reconstruct the left and right image from left and right disparity maps ($ d^{l} $ and $ d^{r} $), the bilinear sampling from the Spatial Transform Networks \cite{jaderberg2015spatial} is used in this paper. The similar approach is also followed in \cite{godard2017unsupervised} for reconstruction from disparity map. The left image is reconstructed from the left disparity map $ d^{l} $ and input right image $ I^{r} $, whereas the right image is reconstructed from the right disparity map $ d^{r} $ and input left image $ I^{l} $ as shown in the Figure \ref{fig:6loss}. The reconstructed left and right images are referred as $ \hat{I}^{l} $ and $ \hat{I}^{r} $ respectively throughout the paper. We also used the loss functions $(C)$ such as appearance matching loss $(C_{ap})$, disparity smoothness loss $(C_{ds})$  and left-right consistency loss $(C_{lr})$ similar to \cite{godard2017unsupervised} but in dual network framework. The loss functions are defined below.

\textbf{Appearance Matching loss:} To enforce the appearance of estimated images must be similar to the input image, a combination of  L1 norm and Structural Similarity Index Metric (SSIM) \cite{wang2004image} loss term is used for both left and right images, defined as \cite{godard2017unsupervised},

\begin{equation}
C^{\beta}_{ap} = \frac{1}{N} \sum_{i, j} \alpha \frac{1 - SSIM(I^{\beta}_{ij}, \hat{I}^{\beta}_{ij})}{2} + (1-\alpha) \parallel{I^{\beta}_{ij} - \hat{I}^{\beta}_{ij}}\parallel
\end{equation}
where $\beta \in \{l, r\}$, $C^{l}_{ap}$  refers appearance matching loss between estimated left image and input left image and $C^{r}_{ap}$ refers appearance matching loss between estimated right image and input right image and $\alpha$ represents the weight between SSIM and L1 norm.

\textbf{Disparity Smoothness Loss:} The image gradient based disparity smoothness loss is computed from both disparity maps to ensure the estimated disparity map should be smooth. Similar to \cite{godard2017unsupervised}, the disparity smoothness loss is given as,

\begin{figure*}[t!]\centering
\includegraphics[width=\textwidth]{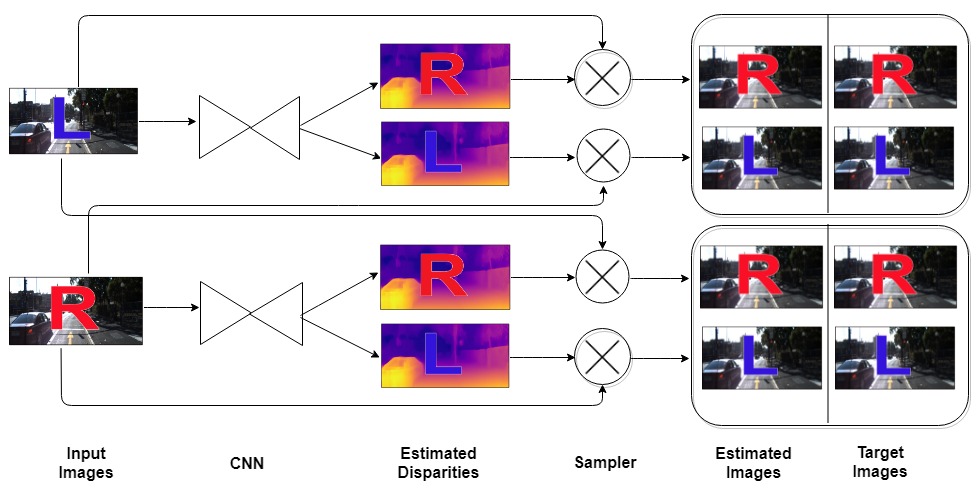}
\caption{Pictorial representation of our Dual Network Model with 12 losses (DNM12)}
\label{fig:12loss}
\end{figure*}

\begin{equation}
C^{\beta}_{ds} = \frac{1}{N}\sum_{i,j}|\partial_{x}d^{\beta}_{ij}| e^{-\parallel\partial_{x}I^{\beta}_{ij}||} + |\partial_{y}d^{\beta}_{ij}|e^{-\parallel\partial_{y}I^{\beta}_{ij}||}
\end{equation}
where $\beta \in \{l, r\}$, $C^{l}_{ds}$  refers the disparity smoothness loss of left disparity map $d^{l}$ estimated by CNN-L, $C^{r}_{ds}$ refers the disparity smoothness loss of right disparity map $d^{r}$ estimated by CNN-R and $\partial$ is the partial derivative.

\textbf{Left Right Consistency Loss:} To maintain the estimated left disparity map $d^{l}$ and right disparity map $d^{r}$ to be consistent, the L1 term penalties on estimated disparities similar to \cite{godard2017unsupervised} are computed between $d^{l}$ and $d^{r}$ as follows,

\begin{equation}
C_{lr} = \frac{1}{N}\sum_{i,j}|d^{l}_{ij} - d^{r}_{ij + d^{l}_{ij}}|
\quad \textrm{and} \quad
C_{rl} = \frac{1}{N}\sum_{i,j}|d^{r}_{ij} - d^{l}_{ij + d^{r}_{ij}}|
\end{equation}
where $C_{lr}$ and $C_{rl}$ refer the left to right and right to left consistency losses respectively. 

Similar to Godard et al. \cite{godard2017unsupervised}, four output scales $s$ in both left and right CNNs are used in this paper in order to make the loss functions more robust. The combined cost function $C_{s}$ at scale $s$ including all above losses i.e. appearance matching losses $C^{l}_{ap}$ and $C^{r}_{ap}$, disparity smoothness losses $C^{l}_{ds}$ and $C^{r}_{ds}$ and left-right consistency losses $C_{lr}$ and $C_{rl}$ is given as $ C_{s} = \alpha_{ap}(C^{l}_{ap} + C^{r}_{ap}) + \alpha_{ds}(C^{l}_{ds} + C^{r}_{ds}) + \alpha_{lr}(C_{lr} + C_{rl})$. The final Cost/Loss function for proposed DNM6 model is computed as $ C = \sum_{s=1}^{4} C_{s} $ at different output scales from s = 1 to 4 similar to \cite{godard2017unsupervised}. At testing time, a single left image, $I^{l}$ is needed as the input to the left CNN (i.e., CNN-L) and it predicts the disparity map $d^{l}$ from the trained network. Note that, the right CNN with input $I^{r}$ can also be used to predict the disparity map $d^{r}$. Once disparity map $d$ (i.e. $d^{l}$ or $d^{r}$) is computed, it can be converted into depth map $(D)$ as $D = \frac{f \times B}{d}$, where $f$ represents the focal length and $B$ is the baseline between stereo cameras.

\subsection{Dual Network Model with 12 Losses (DNM12)}
In our previous DNM6 model, disparity maps are estimated from each network individually, whereas in this DNM12 model, the left-right cross disparity mapping is also proposed as depicted in Figure \ref{fig:12loss}. The left and right CNN networks of DNM6 are extended to generate two output disparities (i.e. left and right) from each CNN. Similar to Godard et al. \cite{godard2017unsupervised}, it generates both left and right disparity maps from a single image. During training, the left image $ I^{l} $ and right image $ I^{r} $ of stereo pair are provided as inputs to the left CNN (CNN-L) and right CNN (CNN-R) respectively. In DNM12 architecture, both the CNN's predict the left and right disparities independently as illustrated in Figure \ref{fig:12loss}. Here, we consider $d^{l_{l}}$ and $d^{l_{r}}$ as the left and right disparity maps respectively estimated by the left CNN-L and similarly $ d^{r_{l}} $ and $ d^{r_{r}} $ as the left and right disparity maps respectively estimated by the right CNN-R. As shown in the Figure \ref{fig:12loss}, four bilinear samplers are used for reconstructing the two output left images $\hat{I}^{l_{l}}$ and $\hat{I}^{r_{l}}$ corresponding to left input image and two output right images $\hat{I}^{l_{r}}$ and $\hat{I}^{r_{r}}$ corresponding to right input image. The $\hat{I}^{l_{l}}$ uses $d^{l_{l}}$ and $I^{r}$, $\hat{I}^{l_{r}}$ uses $d^{l_{r}}$ and $I^{l}$, $\hat{I}^{r_{l}}$ uses $d^{r_{l}}$ and $I^{r}$, and $\hat{I}^{r_{r}}$ uses $d^{r_{r}}$ and $I^{l}$. 
In DNM12, four appearance matching losses, four disparity smoothness losses and four left-right consistency losses are considered. 

The \textit{Four Appearance Matching Losses} are defined as follows,
\begin{equation}
C^{\beta_{\gamma}}_{ap} =  \frac{1}{N} \sum_{i, j} \alpha \frac{1 - SSIM(I^{\gamma}_{ij}, \hat{I}^{\beta_{\gamma}}_{ij})}{2} + (1-\alpha) \parallel{I^{\gamma}_{ij} - \hat{I}^{\beta_{\gamma}}_{ij}}\parallel
\end{equation}
where $\beta \in \{l,r\}$, $\gamma \in \{l,r\}$, $C^{l_l}_{ap}$ and $C^{l_r}_{ap}$ are the appearance matching losses for left CNN-L and $C^{r_l}_{ap}$, $C^{r_r}_{ap}$ are the appearance matching losses for right CNN-R. The total appearance matching loss is given by $C_{ap}=(C^{l_l}_{ap}+C^{l_r}_{ap}+C^{r_l}_{ap}+C^{r_r}_{ap})$.

The \textit{Four Disparity Smoothness Losses} are computed as follows,
\begin{equation}
C^{\beta_\gamma}_{ds}  =  \frac{1}{N}\sum_{i,j}|\partial_{x}d^{\beta_{\gamma}}_{ij}| e^{-\parallel\partial_{x}I^{\beta}_{ij}||} + |\partial_{y}d^{\beta_{\gamma}}_{ij}|e^{-\parallel\partial_{y}I^{\beta}_{ij}||}
\end{equation}
where $\beta \in \{l,r\}$, $\gamma \in \{l,r\}$, $C^{l_l}_{ds}$, $C^{l_r}_{ds}$ are the disparity smoothness losses for left CNN-L and $C^{r_l}_{ds}$, $C^{r_r}_{ds}$ are the disparity smoothness losses for right CNN-R. The total disparity smoothness loss is computed as $C_{ds}=(C^{l_l}_{ds}+C^{l_r}_{ds}+C^{r_l}_{ds}+C^{r_r}_{ds})$.

The \textit{Four Left-Right Consistency Losses} are calculated as follows,
\begin{equation}
C^l_{lr} = \frac{1}{N}\sum_{i,j}|d^{l_{l}}_{ij} - d^{l_{r}}_{ij + d^{l_{l}}_{ij}}|, \quad \textrm{and} \quad
C^{l}_{rl} = \frac{1}{N}\sum_{i,j}|d^{l_{r}}_{ij} - d^{l_{l}}_{ij + d^{l_{r}}_{ij}}|
\end{equation}
\begin{equation}
C^{r}_{lr} = \frac{1}{N}\sum_{i,j}|d^{r_{l}}_{ij} - d^{r_{r}}_{ij + d^{r_{l}}_{ij}}|, \quad \textrm{and} \quad
C^{r}_{rl} = \frac{1}{N}\sum_{i,j}|d^{r_{r}}_{ij} - d^{r_{l}}_{ij + d^{r_{r}}_{ij}}|
\end{equation}
where $C^{l}_{lr}$, $C^{l}_{rl}$ are the left-right and right-left consistency losses for left CNN-L and $C^{r}_{lr}$, $C^{r}_{rl}$ are the left-right and right-left consistency losses for right CNN-R. The total left-right consistency loss is calculated as $C_{lr}=(C^l_{lr}+C^l_{rl}+C^r_{lr}+C^r_{rl})$.

Similar to DNM6, the total Loss function in DNM12 is also defined as $ C = \sum_{s=1}^{4} C_{s} $ at different output scales from s = 1 to 4, where $C_{s}$ at a particular scale is computed by weighted sum of all losses as $C_{s} = \alpha_{ap} \times C_{ap} +  \alpha_{ds} \times C_{ds} + \alpha_{lr} \times C_{lr}.$
The same procedure as provided in previous DNM6 model is followed in DNM12 also for testing, a single image is taken as input to either CNN-L or CNN-R and it predicts the disparity map from the trained network which is converted into depth map.

\begin{table*}[t!]
\caption{Experimental results by using proposed dual CNN based DNM6 and DNM12 models for unsupervised depth estimation over KITTI benchmark database. The training is done over KITTI training images and the evaluation is done over KITTI test images. In this table, pp denotes the post-processing. The best results without post-processing are highlighted in bold face.}
\centering
\begin{tabular}{|>{\small}p{0.25\linewidth}|>{\small}p{0.09\linewidth}|>{\small}p{0.08\linewidth}|>{\small}p{0.08\linewidth}|>{\small}p{0.12\linewidth}|>{\small}p{0.09\linewidth}|>{\small}p{0.07\linewidth}|>{\small}p{0.07\linewidth}|>{\small}p{0.07\linewidth}|}
\hline
& \multicolumn{5}{|c|}{Lower is better} & \multicolumn{3}{|c|}{Higher is better} \\
\cline{2-9}
Method & Abs Rel & Sq Rel & RMSE & RMSE log & d1-all & \textbf{a1} & \textbf{a2} & \textbf{a3} \\
\hline
Godard et al. \cite{godard2017unsupervised} No LR & 0.123 & 1.417 & 6.315 & 0.220 & 30.318 & 0.841 & 0.937 & 0.973 \\
Godard et al. \cite{godard2017unsupervised} & 0.124 & 1.388 & 6.125 & 0.217 & \textbf{30.272} & 0.841 & 0.936 & 0.975 \\
DNM6 Model & 0.1223 & 1.4004 & 6.162 & 0.214 & 31.050 & \textbf{0.848} & \textbf{0.941} & \textbf{0.976}\\
DNM12 Model & \textbf{0.1221} & \textbf{1.3058} & \textbf{6.069} & \textbf{0.213} & 31.455 & 0.841 & 0.939 & \textbf{0.976}\\
\hline
DNM6 Model PP & 0.1157 & 1.2037 & 5.830 & 0.203 & 30.004 & 0.852 & 0.945 & 0.979\\
DNM12 Model PP & 0.1157 & 1.1404 & 5.772 & 0.203 & 30.342 & 0.848 & 0.944 & 0.979\\
\hline
\end{tabular}
\label{result:comparison}
\end{table*}

\section{Experimental Results and Analysis}
We have used the standard datasets such as KITTI and Cityscapes for the experiments. The KITTI database \cite{geiger2012we} consists of stereo pairs from different scenes. Similar to Godard's work \cite{godard2017unsupervised}, 29,000 stereo pairs are used for training and 200 high-quality images are used as the test cases along with its depth maps. The Cityscapes database \cite{cityscapes} contains the stereo pairs captured for autonomous driving. Similar to Godard's work \cite{godard2017unsupervised}, we have used the 22,973 stereo pairs for training after cropping each image such that the 80\% of the height is preserved and the car hoods are removed. Similar to \cite{godard2017unsupervised}, we have used the same 200 KITTI stereo images for testing over Cityscapes database.

The CNN architectures in our network are same as in Godard et al. \cite{godard2017unsupervised}. 
The proposed DNM6 and DNM12 models are implemented in TensorFlow which contains 62 million trainable parameters. 
We have used following parameters, $\alpha = 0.85,$  $\alpha_{ap} = 1,$ $\alpha_{ds} = 0.1,$ $\alpha_{lr} = 1.0 $ and learning rate $\lambda = 10^{-4} $ for first 30 epochs and $0.5 $ x $  10^{-4}$ for next 10 epochs and $0.25 $ x $ 10^{-4}$ for the last 10 epochs. The data augmentation is done on fly, similar to \cite{godard2017unsupervised}. During test time, a post-processing is performed to reduce the effect of stereo dis-occlusions similar to \cite{godard2017unsupervised}.

In both DNM6 and DNM12 methods, the estimated disparity map $d(x)$ is further converted into depth map as $D(x)=\frac{fB}{d(x)}$, where $f$ is the focal length and $B$ is the baseline. The evaluation of both models are done with the estimated depth maps $D(x)$ and provided ground truth depth maps $G(x)$. The evaluation metrics are same as in \cite{godard2017unsupervised} such as  Absolute Relative difference (\textbf{Abs Rel}), Squared Relative difference (\textbf{Sq Rel}), Root Mean Square Error (\textbf{RMSE}), \textbf{RMSE log}, and \textbf{d1-all}.
The lower values of these metrics represent the better performance. We also measured the \textit{Accuracy metrics} (i.e., \textbf{\mbox{a1}}, \textbf{\mbox{a2}}, and \textbf{\mbox{a3}} similar to \cite{godard2017unsupervised}) for which higher is better. 


The results are reported in Table \ref{result:comparison} over KITTI database and compared with very recent state-of-the-art unsupervised method proposed by Godard et al. \cite{godard2017unsupervised} with and without left-right (LR) consistency. Note that the lower values of \textbf{Abs Rel}, \textbf{Sq Rel}, \textbf{RMSE}, \textbf{RMSE log}, and \textbf{d1-all} and the higher values of accuracies \textbf{a1}, \textbf{a2}, and \textbf{a3} represent the better performance. The performance of proposed DNM6 and DNM12 methods are also tested with a pre-procesing (\textbf{PP}) step to reduce the effect of stereo dis-occlusions \cite{godard2017unsupervised}. 
The best results without PP are highlighted in bold face in Table \ref{result:comparison}. It can be easily observed that the proposed dual CNN based models i.e. both DNM6 and DNM12 perform better than Godard et al. \cite{godard2017unsupervised} with and without left-right consistency. The \textbf{Abs Rel}, \textbf{Sq Rel}, \textbf{RMSE}, \textbf{RMSE log}, and \textbf{d1-all} values are generally lower and accuracies \textbf{a1}, \textbf{a2}, and \textbf{a3} are higher for the proposed DNM6 and DNM12 methods. It is also noticed that DNM12 completely outperforms the Godard et al. \cite{godard2017unsupervised} in all terms except \textbf{d1-all}. The performance of DNM6 model is improved in terms of the \textbf{Abs Rel}, \textbf{RMSE}, \textbf{a1}, \textbf{a2}, and \textbf{a3} as compared to the Godard model. The DNM12 model exhibits the better performance as compared to the DNM6 model in all terms except accuracies. As for as accuracies are concerned, the DNM6 model is superior as compared to DNM12 model because generating right disparity from left image and left disparity from right image is not suited for pixel level thresholding. This is also seen that the performance of proposed models improved significantly with post-processing step over KITTI database.

\begin{table*}[t!]
\caption{Experimental results by using proposed dual CNN based DNM6 and DNM12 models for unsupervised depth estimation over Cityscapes benchmark database. The training is done over Cityscapes training images and the evaluation is done over KITTI test images. In this table, pp denotes the post-processing. The best results without post-processing are highlighted in bold face.}
\centering
\begin{tabular}{|>{\small}p{0.25\linewidth}|>{\small}p{0.09\linewidth}|>{\small}p{0.08\linewidth}|>{\small}p{0.08\linewidth}|>{\small}p{0.12\linewidth}|>{\small}p{0.09\linewidth}|>{\small}p{0.07\linewidth}|>{\small}p{0.07\linewidth}|>{\small}p{0.07\linewidth}|}
\hline
& \multicolumn{5}{|c|}{Lower is better} & \multicolumn{3}{|c|}{Higher is better} \\
\cline{2-9}
Method & Abs Rel & Sq Rel & RMSE & RMSE log & d1-all & \textbf{a1} & \textbf{a2} & \textbf{a3} \\
\hline
Godard et al. \cite{godard2017unsupervised} & 0.699 & 10.060 & 14.445 & 0.542 & 94.757 & 0.053 & 0.326 & 0.862 \\
DNM6 Model & 0.2704 & 3.7637 & 9.186 & 0.326 & 64.215 & 0.649 & 0.864 & 0.941\\
DNM12 Model & \textbf{0.2661} & \textbf{3.6491} & \textbf{8.915} & \textbf{0.316} & \textbf{61.163} & \textbf{0.669} & \textbf{0.875} & \textbf{0.946} \\
\hline
DNM6 Model PP & 0.2474 & 2.9781 & 8.406 & 0.300 & 63.780 & 0.663 & 0.881 & 0.954\\
DNM12 Model PP & 0.2396 & 2.8945 & 8.178 & 0.289 & 58.733 & 0.687 & 0.889 & 0.959 \\
\hline
\end{tabular}
\label{result:cityscapes}
\end{table*}

The results comparison of proposed models with Godard et al. \cite{godard2017unsupervised} over Cityscapes database is illustrated in Table \ref{result:cityscapes}. In this Table, the training is performed over Cityscapes database, whereas the test images are same as in KITTI database. It is noticed from this experiment that the proposed models are superior than Godard et al. \cite{godard2017unsupervised} over Cityscapes database in all terms. Moreover, the DNM12 model performs better than DNM6 model. As for as both databases are concerned, the results of proposed models over KITTI database is better than the Cityscapes database. The possible reason can be the difference between the camera calibration between training and testing databases. The similar observations are also made by Godard et al. \cite{godard2017unsupervised}. The post-processing step enhances the performance of proposed DNM6 and DNM12 models over Cityscapes database.



\section{Conclusion}
In this paper, the dual CNN based models DNM6 and DNM12 are presented for unsupervised monocular depth estimation. The dual network models used two different CNNs (CNN-L and CNN-R) for left and right images of training stereo pairs respectively. In DNM6 and DNM12, total 6 and 12 losses are used, respectively. The results are computed over benchmark KITTI and Cityscapes databases and compared with the recent left-right consistency based method. It is observed that the DNM12 outperforms the existing method left-right consistency method. It is also observed that the DNM12 model improves the performance over DNM6 model in most of the cases. The post-processing step further boosts the performance of proposed models.

\section*{Acknowledgement}
This research is supported by Science and Engineering Research Board (SERB), Govt. of India through Project Sanction Number ECR/2017/000082.

\bibliographystyle{splncs04}
\bibliography{Reference}

\end{document}